\documentclass{article}

\PassOptionsToPackage{numbers, compress}{natbib}

\usepackage[main, preprint]{neurips_2026}

\usepackage[utf8]{inputenc}
\usepackage[T1]{fontenc}
\usepackage{hyperref}
\usepackage{url}
\usepackage{booktabs}
\usepackage{amsfonts}
\usepackage{amsmath}
\usepackage{amssymb}
\usepackage{nicefrac}
\usepackage{microtype}
\usepackage{xcolor}
\newcommand{\revred}[1]{#1}
\usepackage{graphicx}
\usepackage{xspace}
\usepackage[flushleft]{threeparttable}
\usepackage{makecell}
\usepackage[ruled,vlined,linesnumbered]{algorithm2e}
\usepackage[textsize=tiny]{todonotes}

\def\method{Wayfinder\xspace}

\title{Wayfinder: Adaptive Resource Routing from Agent Citations}

\author{%
  Miguel Romero\thanks{Corresponding author.} \\
  University of Minnesota \\
  \texttt{romer333@umn.edu} \\
  \And
  George Karypis \\
  University of Minnesota \\
  \texttt{karypis@umn.edu}
}

\begin{document}

\maketitle

\begin{abstract}
To act effectively, agents must gather contextual information across resources such as chats, issue trackers, email and knowledge bases, to answer questions like \emph{what was decided in the last project meeting?} or \emph{where is the FAQ for our internal tooling?} Where this information lives in a user's environment is unknown to the agent, so it re-explores the same resources across consecutive interactions. Common routing approaches represent each resource as a tool and rely on high-level descriptions to generalize to unseen environments. Such descriptions often fail to separate similar or overlapping resources and cannot improve from the agent's subsequent interactions. We introduce \method, an adaptive router that uses the documents cited by the downstream agent as feedback about where relevant information resides. These citations update \method's internal resource representations, which subsequent interactions use to prioritize the resources likely to contain relevant information. We also introduce resource augmentation during offline reinforcement learning, which anonymizes and permutes resource identities to discourage training-time overfitting of which resource holds which topic; our ablations suggest it helps most when few training environments are available. Because no openly available professional environments pair realistic queries with resource-access patterns, we construct and release three controlled testbeds derived from existing datasets: NFCorpus, FEVER and Tech Startup. Compared with supervised, tool-retrieval and contextual bandit baselines under a shared closed-loop protocol, \method gives the best routing when resources contain semantically overlapping content that static descriptions cannot distinguish, while description-based tool retrieval suffices when resources are drawn from distinct corpora.
\end{abstract}

\section{Introduction}

Large Language Models (LLMs) are increasingly judged by their ability to act as agents to accomplish tasks on behalf of users \citep{bandi2025mcp,deng2025swe,tbench_2025}. To complete tasks, however, an agent must do more than reason and act: it requires contextual information about the user's environment. Answering \emph{what was decided in the last project meeting?} or \emph{where do I look up the FAQ for internal LiteLLM?} means seeking evidence across resources such as chats, issue trackers, email, or knowledge bases, each reachable only through a tool that retrieves the documents that specific resource contains (e.g., conversations, tickets, articles) \citep{anthropic2026cowork, anthropic2026interactive}.
Where that grounding information resides is user- or organization-specific and is not known to the agent in advance: an ongoing project may be documented in chat threads in one organization, but captured in email conversations or issue tickets in another. This creates a gap between the agent's prior knowledge and the target environment (i.e., the collection of available resources) in which it operates. Lacking that knowledge, the agent must rediscover the layout of the environment during every new user-agent interaction, re-exploring the same resources across consecutive interactions even after earlier ones revealed where some of that information is kept.

A natural response is to treat each resource together with its access tool as a distinct tool, which frames resource routing as tool selection problem. The distinction is that these tools while having a different resource name often have the same search functionality \texttt{[resource\_name].search(query)$\rightarrow$docs}, and differ only in the documents behind them. Thus, any description used for selection must stand in for contents that cannot be enumerated ahead of deployment. The dominant approaches consume exactly such descriptions: an LLM ranks resources from them (ReSLLM \citep{wang2025resllm}), or a contrastively trained bi-encoder embeds them and scores them by similarity to the query (ToolRetriever \citep{qin2024toolllm}). Both inherit the limits of the description: they cannot separate resources whose contents overlap semantically, and they are fixed at deployment, so what the agent discovers in one interaction cannot inform the next. Reinforcement learning has been applied to tool selection \citep{qian2025toolrl}, but reward computation requires gold tool calls which are not available at deployment. Supervised routers such as Learn2Rank \citep{ergashev2023learning} and RouteRAG \citep{guo2025routerag} share similar limitations requiring labeled queries of the target user's environment but a newly encountered environment supplies neither. What remains missing is a routing method that can adapt to a target environment without labels from it.

User--agent interactions already supply a proxy for these labels: answering a query, the agent cites the documents it relied on, and those citations are grounded evidence of where the information a user needs resides in their environment. We introduce \method, an adaptive router that updates itself from those cited documents. \method maintains a bounded representation of each resource, which we call a \emph{resource sketch}, that subsequent interactions consult to prioritize the resources likely to contain the answer. Adaptation therefore happens online, without target-environment labels and without access to the resources' full contents. Routing is advisory rather than restrictive: the agent remains free to search beyond the recommended resources, so the signal continues to arrive even when the router is wrong. The policy itself is trained offline with Reinforcement Learning \citep{shao2024deepseekmath} on training environments; to keep it from memorizing which resource names carry which topics rather than routing on evidence, we anonymize and permute resource identities during training. Our ablations indicate this augmentation matters most when a limited number of training environments is available.

We evaluate in controlled federated environments spanning intra-corpus partitions (NFCorpus, FEVER) and cross-corpus heterogeneity (Tech Startup), with every router run under the same closed-loop protocol. Where resource descriptions read alike, citation-driven adaptation gives the best routing (Table~\ref{tab:exp1_main}): $F1_r$ of $47.67$ on NFCorpus and $64.72$ on FEVER against $33.03$ and $31.01$ for the strongest baselines, while cutting the agent's search calls by $61.2\%$ and $73.7\%$. On Tech Startup, where five disjoint corpora make each resource recognizable from its description, ToolRetriever, a bi-encoder trained on training-environment labels, leads with $98.86$, and \method is second with $86.60$.

Our contributions are as follows:\footnote{The dataset is released at \url{https://huggingface.co/datasets/miguelrc001/wayfinder}.}
\begin{itemize}
    \item \textbf{Problem formulation:} we formalize the \emph{Adaptive Agent Routing Problem}, in which routing policies must be learned with each resource reachable only as a black-box retriever, weak trajectory-level feedback, and without target-environment supervision, while generalizing across environments.
    \item \textbf{Citation-gated supervision and method:} we introduce a supervision paradigm that updates the routing memory only from documents \emph{cited} by the agent, filtering out the noisy retrievals that dominate early exploration (cited rate $74$--$95\%$, cited-doc accuracy $71$--$73\%$ in our trajectories). We instantiate this paradigm in \method, a zero-shot federated router combining bounded resource sketches with a parameter-efficient LoRA (rank $16$) policy trained via RL.
    \item \textbf{Generalization mechanism:} we introduce \emph{session-randomized resource augmentation} during RL post-training. Resource identifiers are anonymized and re-permuted per session, forcing the policy to read environment memory rather than memorize name--topic associations from training environments. Our ablation (Section~\ref{sec:ablation_a2}) shows this augmentation is most useful when training environments are scarce; when many training environments are available the policy already sees enough variability and the augmentation's benefit diminishes.
    \item \textbf{Benchmark:} To the best of our knowledge professional user environments pairing realistic queries with resource-access patterns are not openly available, and no public evaluation captures routing with historical feedback and cross-environment generalization. We therefore construct and release three controlled testbeds derived from existing natural-language datasets to approximate such environments: \emph{Tech Startup}, a cross-corpus setting ($5$ environments, $5$ resources drawn from $5$ distinct real corpora: emails, chat, issues, wiki, code), and \emph{NFCorpus} \citep{boteva2016} and \emph{FEVER} \citep{Thorne18Fever}, two intra-corpus settings ($18$ environments in which resources partition a single BEIR corpus \citep{thakur2021beir}).
\end{itemize}

\section{Problem Formalization}
\label{sec:formalization}

Given a sequence of user queries $\mathcal{Q} = \langle q_1, q_2, ..., q_T\rangle$ issued sequentially to an agent $\pi$, and a collection of $N$ resources $\mathcal{R} = \{r_1, \dots, r_N\}$ (e.g., chats, issue trackers, email, knowledge bases, etc.) we say that a resource $r_j$ is \emph{relevant} to query $q_t$ iff it contains at least one document $d \in r_j$ that grounds the answer to $q_t$. We refer to the collection of relevant resources as $\bar{\mathcal{R}}_t \subseteq \mathcal{R}$.

Our objective is to define a routing function that, given a query $q_t$ and the history of previous interactions between the user and the agent, identifies the resources relevant to $q_t$ and shares them with the agent $\pi$ to make exploration more efficient. We formalize this as the construction of a router $f_{\textit{route}}$ such that:

\begin{equation}
\label{eq:action_space}
    \hat{\mathcal{R}}_t \;=\; f_{\textit{route}}(q_t,\, \mathcal{R},\, \mathcal{H}_{t-1})
\end{equation}

where $\mathcal{H}_{t-1} = \langle (q_i, \tau_i)\rangle_{i<t}$, is the interaction history before $t$, $\tau_i$ is the trajectory the agent produced at round $i$ and $\hat{\mathcal{R}}_t$ $\subseteq \mathcal{R}$ is the set of predicted resources. In this work, we study this problem under the following constraints:
\begin{enumerate}
    \item \textbf{No access to target-environment supervision.}
    During training, the router is learned from a set of training environments and their associated queries, but the target environment and its queries are not available. This setting reflects practical deployments of agentic systems (e.g., Claude Cowork \citep{anthropic2026cowork}) where models must generalize to previously unseen workspaces without access to user-specific data at training time.
    \item \textbf{Restricted resource access via tools.}
    Resources are accessible only through their corresponding tools or APIs, and their contents cannot be exhaustively enumerated or randomly sampled. Consequently, methods that assume full corpus access or centralized indexing across resources are not applicable in this setting.
\end{enumerate}

These constraints define a learning problem in which routing decisions must be inferred under partial observability and weak supervision, distinguishing it from standard tool use settings where the tool description and functionality does not vary across environments (bash, calculator, git, etc.). Because the router must generalize from training environments to previously unseen target deployments, we refer to this routing problem as the \textbf{Adaptive Agent Routing Problem}.
\section{Related Work}
\label{sec:related_work}

\paragraph{End-to-end RL-based RAG systems.} Recent work optimizes retrieval-augmented generation pipelines end-to-end with structured or verifiable rewards, jointly training query reformulation, retrieval, reranking, or knowledge-graph traversal \citep{chen2025improving,hao2025dynasearcher}. These systems assume direct control over the retrieval stack. \method instead isolates the upstream problem of \emph{resource routing}, treats each resource as a black-box retriever, and acts as a parameter-efficient routing layer that can be composed with such pipelines rather than replacing them.

\paragraph{Resource selection and bandit routing.} Resource selection has a long history in distributed information retrieval and federated search \citep{shokouhi2011federated,balog2012collection}. Classical methods like CORI \citep{Callan2002DISTRIBUTEDIR} use term statistics with no learning. Supervised approaches \citep{guo2025routerag,ergashev2023learning,10.1145/1835449.1835468,10.1145/3077136.3080657} learn classifiers or rankers from labeled query--resource pairs and perform well in data-rich, stationary deployments, but assume stable resource semantics and degrade under distribution shift or in low-data target environments. A complementary line formulates routing as a bandit problem with online reward attribution \citep{auer2002finite,garivier2008upper}. In agentic settings, however, feedback is sparse, trajectory-level, and mediated by multi-step tool use, complicating per-arm credit assignment, and standard bandit formulations maintain only compact per-arm statistics rather than structured interaction context.

\paragraph{Tool retrieval and zero-shot tool generalization.} Treating each resource together with its access tool as a distinct tool makes our action space formally an instance of tool selection with the key distinction that descriptions, when used, have to reflect the content of the resources as the functionality and signature are largely the same across tools (to retrieve documents and  \texttt{search(query)$\rightarrow$docs}). ReSLLM \citep{wang2025resllm} is a method that uses a description of the resource contents to describe the tool which an LLM (used as the router) uses to rank the tools. On the other hand dense-retrieval variants like ToolRetriever \citep{qin2024toolllm} embeds the same descriptions with a contrastively trained bi-encoder and scores them by similarity to the query .

Reinforcement learning has also been applied to tool selection \citep{qian2025toolrl}, with the reward computed by matching the predicted call against a gold tool call. Learning therefore requires labeled data for these tools. A newly encountered target environment, with a new distribution of resources, carries no such annotation: no query is labelled with the resource that holds its answer. \method instead rewards a routing decision using the documents the primary agent cites while answering requiring no labelled data of the target environment.

\paragraph{Memory-augmented agents.} Long-term memory systems for LLMs (MemoryLLM \citep{wang2024memoryllm}, M+ \citep{wang2025mplus}, MemGPT \citep{packer2023memgpt}) and agentic continual-learning systems such as CLIN \citep{majumder2023clin} compress past interactions to retain content or user preferences. These approaches target \emph{content retention} rather than the \emph{navigational} problem of identifying which external resources are likely to be useful for a query.

What separates \method from the work above is its supervision, not its components. Most routers discussed decide from evidence fixed before deployment: a resource description (ReSLLM, ToolRetriever), corpus statistics over enumerated contents (CORI, ReDDE), or labelled query--resource pairs from the target environment (Learn2Rank, RouteRAG). \method instead uses the agent's citations to adapt to the target environment and become more efficient as the user interacts more with the agent.

\section{Methodology: \method}
\label{sec:methodology}

\begin{figure*}
    \centering
    \includegraphics[width=1\linewidth]{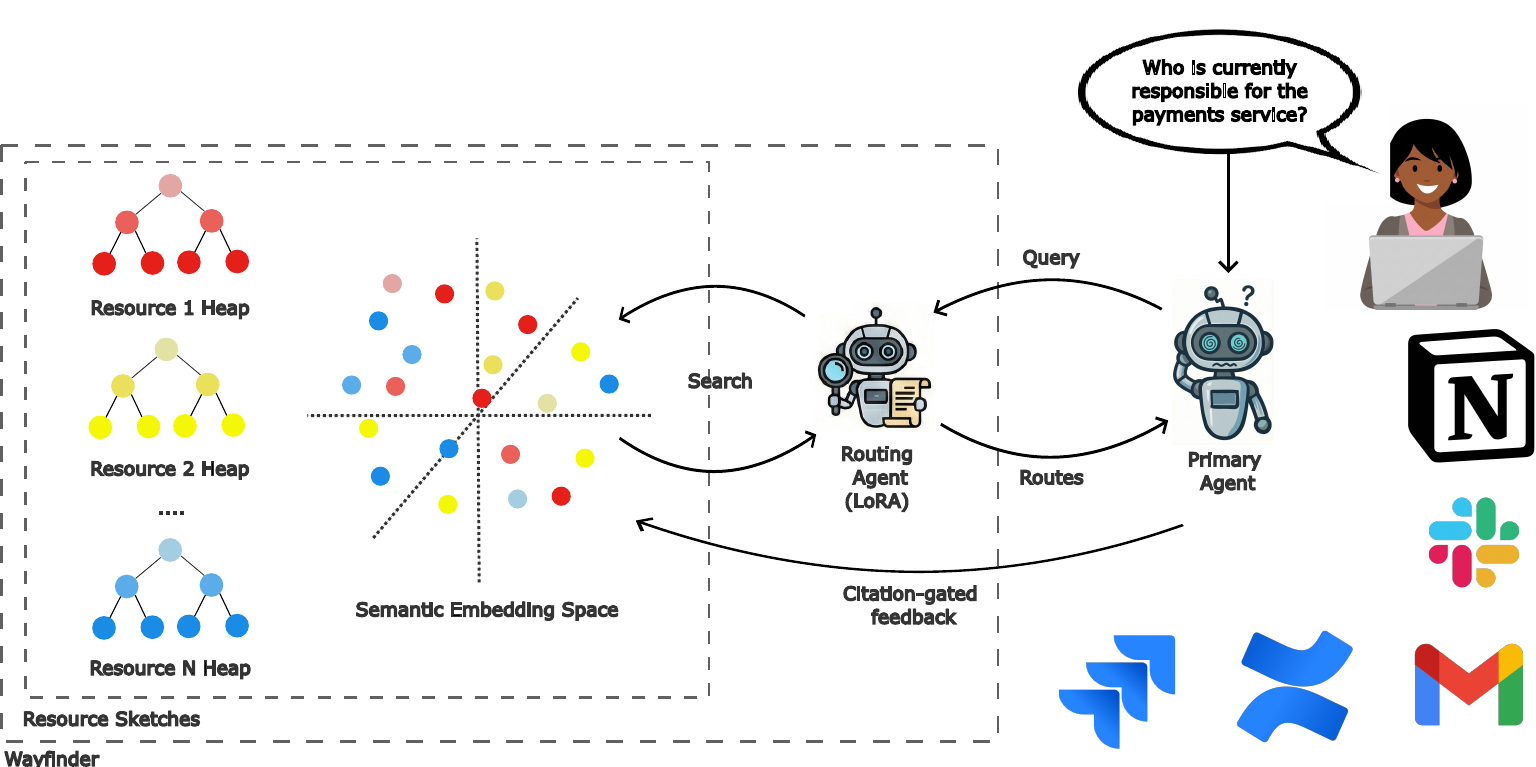}
    \caption{System overview of \method. A LoRA-adapted routing policy interacts with a bounded environment memory of \emph{resource sketches} to select resources for downstream tool invocation, and updates its routing behavior based on citation-derived feedback from the primary agent.}
    \label{fig:method}
\end{figure*}

When we open a new agent session and ask it to perform an action or answer a question the agent often needs to explore the environment to gather context, learning along the way where the information that is relevant to that user is kept. This knowledge is lost when the session ends, and the next session has to repeat the same exploration. We argue that this is inefficient. Queries about the same project tend to be answered from the same few resources, so earlier sessions can inform later ones. Citations also point at evidence the agent actually used to answer a question. \method builds upon this observation and proposes a citation-gating mechanism to retain and utilize this knowledge across sessions.

\method is a routing method that is paired with an agent $\pi$ to make its trajectories more efficient over time. When a new query is passed to $\pi$ it can call \method to retrieve the resource where the information is likely to be found. On the other hand, when a trajectory from the agent $\pi$ is completed, this is passed to \method to store the knowledge. At a high level \method is composed of an environment memory $\mathcal{M}$, containing an internal representation of the environment, and a LLM-based routing policy $\pi_{\textit{route}}$ that uses that memory to provide resource recommendations to the agent $\pi$. Algorithm~\ref{alg:routing_loop} states the loop in full: routing (line~3), the agent's rollout (line~5), citation extraction (lines~8--10) and the sketch update (lines~12--19).

\subsection{Citation gating}
\label{sec:method_citations}

A document is written to the memory $\mathcal{M}$ only if the agent \emph{cited} it and its identifier appears in one of the tool outputs. Formally, given the cited document identifiers $I_t$ and the tool-call log $\{(r_{i,k}, D_{i,k})\}_k$ of the agent's trajectory $\tau_t$, where the $k$-th pair records the resource searched $r_{i,k}$ and the documents it returned $D_{i,k}$, we define the verified citation set $\mathcal{C}_t$ as the documents that are both cited and retrieved:

\begin{equation}
\label{eq:citation_extraction}
\mathcal{C}_t \;=\; \big\{\, d \,:\, \textit{id}(d) \in I_t \;\wedge\; d \in \textstyle\bigcup_k D_{i,k} \,\big\}.
\end{equation}

Identifiers matching no retrieved document are discarded. We write $\rho(\mathcal{C}_t)$ to note the set of resources these documents came from:

\begin{equation}
\label{eq:cited_resources}
\rho(\mathcal{C}_t) \;=\; \big\{\, r_{i,k} \,:\, \exists\, d \in \mathcal{C}_t \cap D_{i,k} \,\big\}.
\end{equation}

If the agent terminates with no cited documents, $I_t = \emptyset$, then $\mathcal{C}_t = \emptyset$ and no update occurs. Restricting $\mathcal{M}$ to documents that were cited and appear in a tool output keeps evidence the agent judged relevant to the user, rather than material it merely explored, and filters out hallucinated documents.

We characterize this signal over a query stream with two statistics: the \emph{citation rate} $\Pr_t[\mathcal{C}_t \neq \emptyset]$, how often citations are available, and the \emph{citation accuracy} $\mathbb{E}_t[\,|\rho(\mathcal{C}_t) \cap \bar{\mathcal{R}}_t| / |\rho(\mathcal{C}_t)| \mid \mathcal{C}_t \neq \emptyset\,]$, how often they name the right resource given that a citation exists, where $\bar{\mathcal{R}}_t$ is the ground-truth resource set for $q_t$.

\subsection{Resource Sketches}
\label{sec:method_loop}

Evidence is stored per resource. For each $r_j \in \mathcal{R}$ the \emph{resource sketch} $\mathcal{M}^j$ is a bounded set of at most $M$ records, each a triple $(d, q, t)$: a cited document $d$ observed via a tool call to $r_j$, the query $q$ that elicited the citation, and an access timestamp. The \emph{environment memory} is then the collection of these resource sketches $\mathcal{M} = (\mathcal{M}^1, \dots, \mathcal{M}^N)$. Sketches start empty, $\mathcal{M}_0^j = \emptyset$, so on the first query the policy may emit an empty action set and let the agent fall back on static descriptions. Thus, when \method is applied to a new environment it starts with an empty memory but a trained routing policy $\pi_{\textit{route}}$.

On each rollout, every document in $\mathcal{C}_t$ retrieved from a cited resource $r_j \in \rho(\mathcal{C}_t)$ is inserted into $\mathcal{M}^j$ as a record $(d, q_t, t)$, evicting the least-recently-used record when the sketch is at capacity $M$. Resources outside $\rho(\mathcal{C}_t)$ are untouched. Three consequences follow: retrieved-but-uncited documents never enter, so early exploration does not pollute sketches with low-precision associations; when the agent fails to cite, sketches go stale rather than wrong, giving slower but stable adaptation; and the capacity $M$ caps the storage and retrieval cost of the routing layer regardless of stream length. Lastly, a centralized semantic index over $\mathcal{M}$ supports retrieval across resource sketches of the top-$k$ documents relevant to a query.

\begin{algorithm}[t]
\caption{\method routing loop and citation-gated memory update}
\label{alg:routing_loop}
\small
\KwIn{Resources $\mathcal{R}$; query stream $\{q_t\}_{t=1}^T$; routing policy $\pi_{\textit{route}}$; primary agent $\pi$; budget $B$; per-resource sketch capacity $M$.}
\KwOut{Routing decisions $\{\hat{\mathcal{R}}_t\}_{t=1}^T$ and final environment memory $\mathcal{M}_T$.}
\BlankLine
Initialize $\mathcal{M}_0 \leftarrow (\emptyset, \dots, \emptyset)$\;
\For{$t = 1, \dots, T$}{
    \tcp{(1) Routing}
    $\hat{\mathcal{R}}_t \leftarrow \textsc{Route}(\pi_{\textit{route}}, q_t, \mathcal{M}_{t-1}, \mathcal{R})$ \tcp*{$|\hat{\mathcal{R}}_t| \le B$}
    \BlankLine
    \tcp{(2) Primary-agent rollout and parsing}
    $\tau_t \leftarrow \pi(q_t, \hat{\mathcal{R}}_t)$\;
    $\big(y_t,\, \{(r_{i,k}, D_{i,k})\}_k\big) \leftarrow \textsc{ParseTrajectory}(\tau_t)$\;
    \BlankLine
    \tcp{(3) Citation extraction (Eqs.~\ref{eq:citation_extraction}--\ref{eq:cited_resources})}
    $I_t \leftarrow$ identifiers of the documents cited in $\tau_t$\;
    $\mathcal{C}_t \leftarrow \{ d : \textit{id}(d) \in I_t \wedge d \in \bigcup_k D_{i,k} \}$\;
    $\rho(\mathcal{C}_t) \leftarrow \{ r_{i,k} : \exists\, d \in \mathcal{C}_t \cap D_{i,k} \}$\;
    \BlankLine
    \tcp{(4) Sketch update (citation-gated, LRU-bounded)}
    $\mathcal{M}_t \leftarrow \mathcal{M}_{t-1}$\;
    \If{$\mathcal{C}_t \neq \emptyset$}{
        \ForEach{$r_j \in \rho(\mathcal{C}_t)$}{
            \ForEach{$d \in \mathcal{C}_t$ retrieved from $r_j$}{
                \If{$|\mathcal{M}_t^j| \ge M$}{Evict LRU record from $\mathcal{M}_t^j$\;}
                $\mathcal{M}_t^j \leftarrow \mathcal{M}_t^j \cup \{(d, q_t, t)\}$\;
            }
        }
        Update centralized semantic index over $\mathcal{M}_t$\;
    }
}
\end{algorithm}

\subsection{Routing}
\label{sec:meta_training}

Given $q_t$, the routing policy $\pi_{\textit{route}}$ reads the environment memory state at time $t-1$, $\mathcal{M}_{t-1}$, through the semantic index in a ReAct-style loop, optionally issuing further memory searches with refined queries and reasoning over what it retrieves before emitting $\hat{\mathcal{R}}_t$. The agent then executes $\tau_t = \pi(q_t, \hat{\mathcal{R}}_t)$ and the trajectory returns for citation extraction, closing the loop.

The routing policy is implemented as a LLM trained with reinforcement learning with verifiable rewards \citep{lambert2024tulu} and GRPO \citep{shao2024deepseekmath} on training environments only. Specifically, for each query we sample a group of $G$ rollouts and score rollout $i$ by $R_i$ $= F_1(\hat{\mathcal{R}}_t^{(i)}, \bar{\mathcal{R}}_t)$ against the ground-truth set $\bar{\mathcal{R}}_t = \{r_j \in \mathcal{R} : r_j \cap D_t \neq \emptyset\}$, where $D_t$ is the set of relevance-judged documents for $q_t$; malformed output (invalid identifiers or non-parseable JSON) scores zero. The $G$ scores are then centred and scaled against each other, $A_i = (R_i - \mu_q)/(\sigma_q + \epsilon)$, where $\mu_q$ and $\sigma_q$ are the mean and standard deviation of $\{R_i\}_{i=1}^{G}$ and $\epsilon>0$ guards the case where all $G$ rollouts score alike; a rollout that beats its siblings on the same query receives positive weight, one that trails them negative. The policy is then updated by advantage-weighted cross-entropy \citep{williams1992simple} over the router's own generated tokens.

Labeled data $D_t$ is only available on training environments; at testing time \method is applied to a new environment with unseen resources and adapts from citations alone.

\subsection{Forcing the policy to read evidence, not identifiers}
\label{sec:method_aug}

A policy trained on a small number of environments can memorize which identifier carries which topic (``coding questions go to resource A''). The shortcut scores well in training and collapses in a target environment for which resources may carry different content, so it has to be prevented during training rather than detected afterwards. We therefore apply two augmentations while training the policy, both at the granularity of a \emph{session}: a single training rollout, within which one assignment of identifiers to resources is held fixed.

\emph{Resource encoding.} Descriptive names (``email'', ``chat'') are replaced by anonymous identifiers $\texttt{resource\_1},\dots,\texttt{resource\_N}$ in both the input and the action vocabulary, removing name-based priors.

\emph{Resource randomization.} Each rollout draws a fresh permutation mapping resources to identifiers, applied consistently within that rollout.

Together the two leave no stable association between an identifier and a topic across query rollouts. Sketch content is then the only signal that survives from one rollout to the next, which is the signal that the policy must learn to interpret.

\section{Dataset}
\label{sec:dataset}

Routing in real deployments combines two sources of heterogeneity: \emph{intra-corpus partitioning}, where different organizations split similar content across different resources (e.g., release notes in the wiki vs.\ chat archives); and \emph{cross-corpus heterogeneity}, where different resource types serve different communicative functions and therefore expose different vocabulary, length, and structural style (e.g., syntactic code blocks vs.\ conversational messages). Existing federated-search benchmarks \citep{wang2024feb4rag,10.1145/2740908.2742755} contain too few environments to support train--deploy splits, and BEIR \citep{thakur2021beir} provides single-corpus relevance judgments without independent environments. We therefore evaluate \method on two complementary regimes (Table~\ref{tab:dataset_stats}): \emph{intra-corpus} environments derived from BEIR, and a new \emph{cross-corpus} \emph{Tech Startup} benchmark in which each resource is drawn from a distinct real corpus.

\begin{table}
    \centering
    \begin{threeparttable}
        \caption{Benchmark statistics.}
        \label{tab:dataset_stats}
        \small
        \begin{tabular}{lccccl}
            \toprule
            Dataset& Type & \# Env. & \# Res. /Env. &
            \makecell[c]{Train / Test\\Queries}  & \makecell[c]{Doc. \\ / Res.} \\
            \midrule
            NFCorpus & Intra-corpus & 2 & 50 & 89 / 30 & 121 \\
            Fever & Intra-corpus & 16 & 50 & 333.7 / 82.9 & 6831.8 \\
            Tech Startup & Cross-corpus & 5 & 5 & 180.4 / 45.0 & 150.5 \\
            \bottomrule
        \end{tabular}
        \begin{tablenotes}
            \small
            \item \emph{Intra-corpus} environments partition a single BEIR corpus into semantically coherent resources via GMM clustering. \emph{Cross-corpus} environments draw each resource from a distinct real corpus, yielding heterogeneity at the resource level rather than within a shared corpus. Doc./Res. is the average number of documents per resource; Train/Test follows an $80/20$ temporal split within each environment.
        \end{tablenotes}
    \end{threeparttable}
\end{table}

We derive $18$ environments by partitioning two BEIR corpora into disjoint resources via sentence-transformer embeddings and a Gaussian Mixture Model: the specialized NFCorpus \citep{boteva2016} (2 environments) and the larger fact-verification FEVER corpus \citep{Thorne18Fever} (16 environments). Each environment inherits human-annotated document-level relevance judgments from the source dataset; queries are arranged in an $80/20$ temporal split.

We additionally construct a $5$-environment \emph{Tech Startup} benchmark in which the five resources (emails, chat channels, issue tracker, wiki pages, code repository) are sourced respectively from Enron emails \citep{Klimt2004enron}, the Alpaca instruction-tuning corpus \citep{taori2023alpaca}, GitHub issues from SWE-bench \citep{jimenez2024swebench}, Wikipedia paragraphs from HotpotQA \citep{yang2018hotpotqa}, and source code from CodeSearchNet \citep{husain2019codesearchnet}. Documents and queries are disjoint across environments. Queries are rewritten to reference content rather than the originating application and every retained query passes a semantic-retrieval quality gate (top-5 dense retrieval). Unlike the intra-corpus environments, resource boundaries here reflect natural cross-application heterogeneity rather than synthetic clustering. Full construction and validation details are in Appendix~\ref{app:data}.

\section{Experimental Setup}
\label{sec:exp_setup}

We report three metrics, each a percentage for which higher is better. The router's $F1_{route}$ \citep{shokouhi2011federated,balog2012collection} which scores the accuracy of the recommendation on its own. The agent's citation $F1_{cite}$, which measures the effect of router on the agent's grounding. We do not report answer accuracy: the primary agent can answer many queries in these public benchmarks from parametric knowledge, so a correct answer does not show that relevant evidence was retrieved, whereas $F1_{cite}$ credits only answers grounded in the judged documents. The exploration savings $S$ which measures that effect of the router on the agent's search effort: writing $s(q)$ and $s_0(q)$ for the number of search calls the agent issues on query $q$ with and without a router, then $S = 100\times\mathrm{mean}_q\!\left(1-s(q)/s_0(q)\right)$ is the saving on a typical query.

We compare \method against baselines spanning classical, supervised, tool-retrieval and bandit families. \emph{CORI} \citep{Callan2002DISTRIBUTEDIR} ranks resources from term statistics without learning. \emph{ResLLM} \citep{wang2025resllm} prompts an LLM with document-derived resource summaries. \emph{ToolRetriever} \citep{qin2024toolllm} embeds the same summaries with a bi-encoder contrastively trained on training environments and scores them against the query. \emph{LinUCB-D} \citep{li2010contextual} treats each resource as a bandit arm updated from citation-derived binary feedback. \emph{Learn2Rank} \citep{ergashev2023learning} trains a supervised classifier using labels from the target environment which violates our problem assumptions but we add for reference. LinUCB-D and \method adapt during the stream; CORI, ResLLM and ToolRetriever are fixed once the environment is entered. All embedding-based methods share the same MiniLM-L6-v2 encoder.

\method routing policy is implemented by finetuning Qwen3-8B using  LoRA (rank $16$, $\sim$0.1\% of base parameters) and resource randomization. The routing agent retrieves the top $25$ memory entries per decision step via semantic search to bound its context. Unless otherwise stated, the per-resource sketch capacity is $M=100$. Training runs $125$ GRPO iterations; each iteration samples $8$ queries and generates $8$ rollouts per query (GRPO group size for advantage normalization).

Reported results are average $k$-fold performance where datasets are  partitioned at the environment level ($k\!=\!2$ for NFCorpus, $k\!=\!4$ for FEVER and Tech Startup). Methods like ToolRetriever and \method requiring cross-environment training are trained on the training split. If the method relies on interaction statistics (e.g.,  CORI or \method) those are produced as test environments are evaluated. \method is evaluated closed-loop: its memory is seeded with the documents cited in the target environment's training-split trajectories, without relevance labels, and is then updated from the agent's citations as test queries are processed. As a primary agent we use GPT-4o-mini.

\section{Results}
\label{sec:results}

In this section we present experimental results and ablations under stationary conditions. Due to space constraints the evaluations in non-stationary settings and case study are included in Appendix \ref{sec:exp2_drift}  and Appendix \ref{sec:case_study}, respectively. 

\subsection{Experiment 1: Routers Comparison}
\label{sec:exp1_main}

Our hypothesis is
that citation-derived adaptation pays in proportion to how poorly static descriptions separate
resources. On the intra-corpus benchmarks, roughly fifty resources are drawn from a single
corpus and their summaries read alike, so a router fixed at deployment has little left to
discriminate on; on Tech Startup each resource is a distinct corpus and its description is close
to sufficient on its own. Every router therefore enters the target environment cold and sees the
identical query stream: the adaptive routers revise their beliefs as that stream is processed
while the static ones remain fixed, and each router's recommendations are logged as they are issued and
scored afterwards against the same relevance judgments. \method is evaluated closed-loop: its memory is updated with the citations the primary agent emits after acting on \method's own recommendations, on the same query stream every baseline receives.

\begin{table}
\centering
\begin{threeparttable}
\caption{Closed-loop routing comparison.}
\label{tab:exp1_main}
\small
\setlength{\tabcolsep}{3pt}
\begin{tabular}{lccccccccc}
 & \multicolumn{3}{c}{NFCorpus (2-fold)}& \multicolumn{3}{c}{Fever (4-fold)}& \multicolumn{3}{c}{Tech Startup (4-fold)}\\
\toprule
\textbf{Router} & $F1_{r} \uparrow$ & $F1_{c} \uparrow$ & Saved \% & $F1_{r} \uparrow$ & $F1_{c} \uparrow$ & Saved \% & $F1_{r} \uparrow$ & $F1_{c} \uparrow$ & Saved \% \\
\midrule
\midrule
No routing & -- & $14.42$ & $0.0$ & -- & $33.02$ & $0.0$ & -- & $87.44$ & $0.0$ \\
\midrule
CORI & $29.32$ & $\textbf{20.27}$ & $58.4$ & $7.90$ & $15.83$ & $40.6$ & $66.22$ & $57.78$ & $55.3$ \\
ResLLM & $\underline{33.03}$ & $11.51$ & $63.5$ & $11.77$ & $20.90$ & $45.8$ & $30.16$ & $24.18$ & $22.8$ \\
ToolRetriever & $22.71$ & $13.91$ & $50.4$ & $30.38$ & $\textbf{55.52}$ & $37.9$ & $\textbf{98.86}$ & $\textbf{85.31}$ & $70.4$ \\
LinUCB-D & $26.47$ & $15.43$ & $49.5$ & $9.57$ & $18.33$ & $-4.7$ & $71.78$ & $62.81$ & $58.7$ \\
Learn2Rank & $12.78$ & $5.64$ & $41.1$ & $\underline{31.01}$ & $41.64$ & $9.9$ & $83.28$ & $72.16$ & $59.8$ \\
\method & $\textbf{47.67}$ & $\underline{17.10}$ & $61.2$ & $\textbf{64.72}$ & $\underline{50.88}$ & $73.7$ & $\underline{86.60}$ & $\underline{80.78}$ & $60.4$ \\
\bottomrule
\end{tabular}
\begin{tablenotes}
\small
\item $F1_{r}$: routing F1. $F1_{c}$: the agent's citation F1 against the relevance judgments. Saved\,\%: reduction in the agent's search calls relative to no routing. Bold: best; underline: second best.
\end{tablenotes}
\end{threeparttable}
\end{table}

Routing budgets are $B{=}5$ on NFCorpus and FEVER and $B{=}1$ on Tech Startup. \method uses the same LoRA policy with no retraining and one adapter per fold, and FEVER is capped at $50$ queries per test environment. Each replay is paired with a no-router control on the same queries, which issues $25.90$, $32.90$ and $5.86$ search calls per query on the three datasets. $F1_{c}$ and Saved\,\% are measured by hard-routed replay because the closed loop itself is \emph{advisory}: the agent may search outside the recommendation, so its raw search counts exceed an unrouted agent's and the savings metric is negative by construction, which measures the protocol rather than the router. NFCorpus rests on $n{=}42$: its two test splits yield only $9$ and $33$ GT-usable queries. $F1_{route}$ uses the same RDR-based definition throughout the table; plain set recall would read $28.16$ on NFCorpus, where $43\%$ of queries emit fewer resources than there are gold ones.

On the intra-corpus benchmarks, where roughly fifty resources are drawn from one corpus, \method gives the best routing: $F1_{r}$ of $47.67$ on NFCorpus and $64.72$ on FEVER, against $33.03$ (ResLLM) and $31.01$ (Learn2Rank) for the strongest baselines. FEVER is the dataset where routing is closest to degenerate, since a constant router that ignores the query and selects the five largest resources scores $33.6$ there. Better routing carries through to the agent: \method cuts search calls on a typical query by $61.2\%$ and $73.7\%$ while raising $F1_{c}$ over the unrestricted agent by $2.7$ and $17.9$ points. It does not ground the most answers, however: CORI on NFCorpus ($20.27$) and ToolRetriever on FEVER ($55.52$) reach higher $F1_{c}$. Saved\,\% must be read together with $F1_{c}$, because a router that sends the agent nowhere useful makes it give up early and so \emph{appears} to save exploration: ResLLM saves $63.5\%$ on NFCorpus while its $F1_{c}$ falls below the unrestricted agent's ($11.51$ vs.\ $14.42$).

On Tech Startup, where each resource is a distinct corpus and its description is close to sufficient on its own, ToolRetriever is best on all three metrics ($98.86$, $85.31$ and $70.4\%$). \method is second on routing and citation F1 ($86.60$ and $80.78$), ahead of Learn2Rank ($83.28$), which is trained on target-environment labels, and of LinUCB-D ($71.78$), the other router that adapts online. No router improves on letting the agent search all five resources ($F1_{c}$ of $87.44$). LinUCB-D trails \method on routing on all three datasets and on FEVER increases the agent's search calls ($-4.7\%$ saved).

These results match the two regimes of our hypothesis: citation-derived adaptation gives the best routing where resource descriptions read alike, while description-based tool retrieval suffices where resources are distinct corpora. Although \method employs a reasoning-based routing policy with environment memory, the routing behavior observed in our experiments is consistently shallow. Across all three datasets, trajectories exhibit a median of two routing decisions per query, with over $99\%$ of trajectories completing within two steps (one retrieval step and one action step) and no evidence of long-horizon routing.

\subsection{Ablation 1: RL post-training}
\label{sec:ablation_a1}

We isolate the contribution of RL-based fine-tuning by comparing \method with and without RL post-training while holding memory structure and resource representations constant.

\begin{table}[h]
    \centering
    \begin{threeparttable}
        \caption{Ablation on RL post-training}
        \label{tab:ablation_finetuning}
        \small
        \begin{tabular}{lcccccc}
         & \multicolumn{3}{c}{NFCorpus} & \multicolumn{3}{c}{FEVER} \\
        \toprule
        \textbf{Router} & $F1_{r}$ $\uparrow$ & $F1_{c}$ $\uparrow$ & Saved \% & $F1_{r}$ $\uparrow$ & $F1_{c}$ $\uparrow$ & Saved \% \\
        \midrule
        w/o RL & $41.28$ & $17.77$ & $63.9$ & $66.06$ & $75.93$ & $81.8$ \\
        w/ RL & $\textbf{50.69}$ & $20.05$ & $64.8$ & $\textbf{67.94}$ & $59.70$ & $79.3$ \\
        \bottomrule
        \end{tabular}
        \begin{tablenotes}
            \small
            \item $F1_{r}$: routing F1. $F1_{c}$: the agent's citation F1 against the relevance judgments. Saved\,\%: reduction in the agent's search calls relative to no routing. Bold: higher $F1_{r}$.
        \end{tablenotes}
    \end{threeparttable}
\end{table}

Table~\ref{tab:ablation_finetuning} shows RL-based post-training is critical: removing it consistently degrades routing \revred{$F1_{r}$} across datasets despite identical memory and resource representations, indicating that static inference over environment memory is insufficient and explicit credit assignment is required to optimize routing decisions for downstream retrieval utility. On FEVER, however, the variant without RL grounds \emph{more} answers than the full model ($75.93$ vs.\ $59.70$) while also searching less, so RL post-training buys routing accuracy rather than downstream grounding there.

\subsection{Ablation 2: Resource Augmentation}
\label{sec:ablation_a2}

We evaluate the impact of resource augmentation (Section~\ref{sec:method_aug}) on cross-environment transfer.

\begin{table}[h]
    \centering
    \begin{threeparttable}
        \caption{Ablation on resource augmentation.}
        \label{tab:ablation_augmentation}
        \small
        \begin{tabular}{lcccccc}
         & \multicolumn{3}{c}{NFCorpus} & \multicolumn{3}{c}{FEVER} \\
        \toprule
        \textbf{Router} & $F1_{r}$ $\uparrow$ & $F1_{c}$ $\uparrow$ & Saved \% & $F1_{r}$ $\uparrow$ & $F1_{c}$ $\uparrow$ & Saved \% \\
        \midrule
        w/o res. aug. & $45.35$ & $17.54$ & $56.9$ & $\textbf{68.40}$ & $58.37$ & $76.8$ \\
        w/ res. aug. & $\textbf{50.69}$ & $20.05$ & $64.8$ & $67.94$ & $59.70$ & $79.3$ \\
        \bottomrule
        \end{tabular}
        \begin{tablenotes}
            \small
            \item $F1_{r}$: routing F1 at budget $B{=}5$ (Appendix~\ref{app:metrics}). $F1_{c}$: the agent's citation F1 against the relevance judgments. Saved\,\%: reduction in the agent's search calls relative to no routing. Bold: higher $F1_{r}$.
        \end{tablenotes}
    \end{threeparttable}
\end{table}

Table~\ref{tab:ablation_augmentation} shows resource augmentation improves routing robustness on NFCorpus and slightly degrades performance on FEVER, supporting our hypothesis that augmentation is most beneficial when training environments are scarce and added variability helps prevent overfitting.

\section{Limitations and future work}
\label{sec:limitations}

\method depends on the primary agent's citations. Missing citations slow adaptation, but incorrect ones write wrong evidence into the resource sketches, and routing degrades as their rate grows (Appendix~\ref{app:citation_noise}); weighting updates by citation confidence, decaying unconfirmed evidence, or revising a sketch when later citations contradict it could limit this, and we leave their evaluation to future work. Our benchmarks are controlled testbeds built from public corpora, so they do not capture enterprise permission structures or production-scale latency. Evaluating \method with authentic tools and users is the natural next step.

\section{Conclusion}
\label{sec:conclusions}

We introduced \method, a router that adapts to a target environment from the citations its primary agent emits, without target-environment labels. Under a closed-loop protocol shared by all routers, \method gives the best routing on the intra-corpus benchmarks ($47.67$ and $64.72$ $F1_r$ against $33.03$ and $31.01$ for the strongest baselines) and cuts the agent's search calls by $61.2\%$ and $73.7\%$ while improving its citation F1 over an unrouted agent. Where resources are distinct corpora, as in Tech Startup, description-based tool retrieval suffices and ToolRetriever leads, with \method second. Closed-loop routing stays within $3.3$ points of the offline protocol, and RL post-training improves routing F1 on both intra-corpus benchmarks.

\bibliographystyle{plainnat}
\bibliography{sections/bibliography}

\appendix
\section{Robustness to Non-Stationarity}
\label{sec:exp2_drift}

Non-stationarity is a defining characteristic of federated environments, where the information within resources can change over time independently of the agent’s actions. 
To evaluate robustness to such exogenous changes, we progressively relocate an increasing fraction of ground-truth documents associated with test queries to different resources before testing, simulating changes in where information resides after deployment. 
Higher drift levels subsume lower ones to ensure consistency across settings, and all methods are evaluated under identical drift realizations to isolate differences in routing behavior. 
We report routing recall and precision to assess how effectively each method maintains routing quality as the environment diverges from the training distribution.

\begin{figure}[h]
    \centering
    \includegraphics[width=1\linewidth]{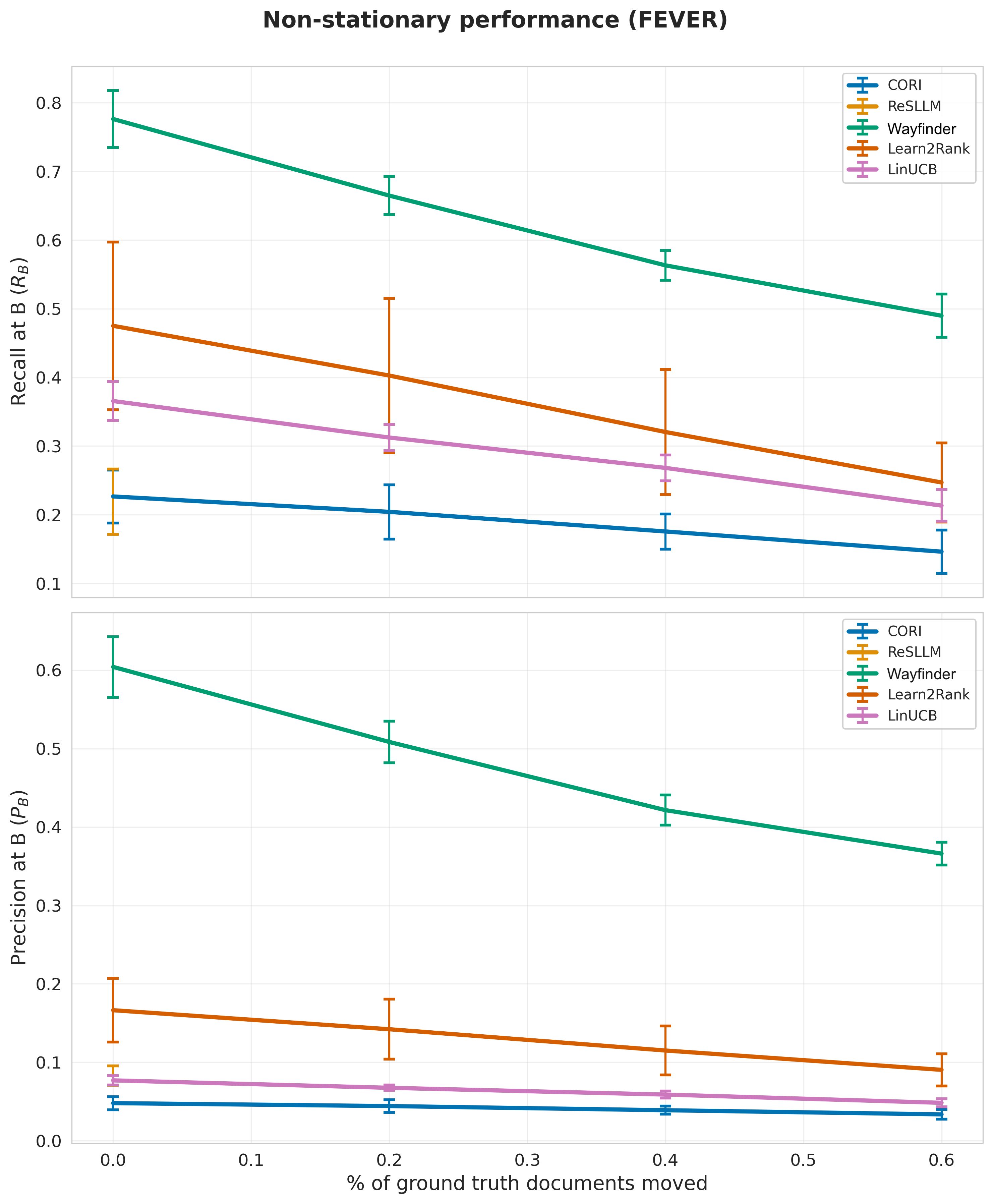}
    \caption{\small Routing robustness under increasing levels of exogenous changes measured by Precision at $B$ and Recall at $B$ with $B{=}5$. As a larger fraction of ground-truth documents is relocated across resources at test time, all methods degrade, with \method exhibiting better adaptation due to the agent feedback loop.}
    \label{fig:non-stationarity}
\end{figure}

Figure~\ref{fig:non-stationarity} shows that routing performance degrades for all methods as the degree of ground-truth movement increases, reflecting the difficulty of routing under severe distribution shift. 
However, \method maintains higher recall and precision across all drift levels, indicating greater robustness to changes in where relevant information resides. 
In contrast, baselines that rely on static assumptions or fixed ranking signals exhibit steeper performance declines as drift increases. 
These results suggest that learning routing policies from interaction feedback yields representations that are less tightly coupled to specific resource assignments, improving resilience to post-deployment environmental changes.

\section{Appendix: Case Study}
\label{sec:case_study}

We present a qualitative case study illustrating how \method performs resource routing using incomplete environment memory, a common setting in federated environments where memory provides only sparse and noisy evidence about resource contents. 

\paragraph{Query.}
\begin{quote}
As of 2014, the electric chair is an optional form of execution in Alabama, Florida, South Carolina, and Idaho.
\end{quote}

Given the query, the agent retrieves semantically related memory samples from multiple resources. Each resource contains a small subset of documents that only partially reflect its underlying collection. \newline

\begin{verbatim}
Resource 34:
  - Hill v. McDonough (U.S. Supreme Court case on
    execution methods in Florida)
  - Capital punishment (overview of death penalty
    practices)
  - Capital punishment debate in the United States
  - List of executions by lethal injection

Resource 51:
  - Shapiro v. McManus (U.S. Supreme Court case on
    judicial procedure)
  - Jordan Sekulow (U.S. legal commentator)
... # omitted for brevity

\end{verbatim}

\begin{verbatim}
<think> Okay, let's tackle this. The user is asking about 
the electric chair being an optional form of execution in 
specific states as of 2014. First, I need to figure out 
which resources might have this information. The 
search_past_interactions tool was called with the query
"electric chair execution states 2014", and the response 
included several documents.

Looking at the documents from resource 34, there's a 
mention of \"Hill v. McDonough,\" which is a Supreme 
Court case about lethal injection in Florida. 
That's related to execution methods. Also, \"Capital 
punishment\" and \"Capital punishment in Bulgaria\"
are in the list, but Bulgaria isn't one of the states 
mentioned. However, the user's query includes Florida, 
so maybe the Hill v. McDonough case is relevant. 

Other resources like 51 have \"Shapiro v. McManus\"
... # omitted for brevity

Since the search results don't have exact matches, but the 
user's query is about execution methods in specific states, 
the most relevant resource might be 34, which includes 
capital punishment-related documents. Even though it 
doesn't directly answer the question, it's the closest. 
... # omitted for brevity

Wait, the user's query is about 2014, and the 
Hill v. McDonough case was in 2015. So that's slightly 
after the user's timeframe. But maybe the information about 
Florida's execution methods is still relevant. Since the 
search results don't have exact matches, but the user needs 
an answer, I should select the resource that has the most 
related documents, even if it's not perfect. Resource 34 
seems the best fit here.</think>
\end{verbatim}
\vspace{4pt}
Although none of the retrieved memory samples explicitly state the queried fact about the electric chair, Resource $34$, the ground truth resource, exhibits a consistent thematic focus on capital punishment and execution methods, including legal cases and policy discussions relevant to U.S.\ states. 
By prioritizing resources with conceptually related topics, \method is able to prioritize relevant resources for the primary agent. This example demonstrates how \method performs robust routing under incomplete memory by inferring where relevant information is likely to reside, rather than relying on exact matches or exhaustive search.

\section{Appendix: Data Construction Process}
\label{app:data}

We construct synthetic federated retrieval environments from BEIR benchmarks to enable controlled evaluation of routing under non-stationarity. Starting from a standard BEIR dataset (corpus, queries, and relevance judgments), we transform static collections into dynamic, multi-resource environments with temporal structure and exogenous drift.

For datasets with graded relevance (e.g., NFCorpus), we retain only highly relevant documents and normalize relevance to binary. Queries without relevant documents after filtering are removed to ensure well-defined routing supervision.

Each dataset is partitioned into multiple independent environments, representing distinct information spaces. Documents are disjointly assigned to environments. Within each environment, documents are organized into a fixed number of resources. Resources are constructed by clustering document embeddings (using a sentence transformer) via a Gaussian Mixture Model, ensuring semantic coherence within each resource and mimic similar content across resources.

To support description-based baselines, each resource is associated with an automatically generated natural-language description. Descriptions are produced by prompting a large language model with representative document samples from the resource (100).

Within each environment, queries are temporally ordered in a query stream to form a single timeline, simulating interactions in a shared environment.

Queries are split temporally into training and test segments using an 80/20 ratio. When enabled, exogenous drift events are realized by randomly moving documents across resources. Drifts are applied only during the test period and preferentially target ground-truth relevant documents to mimic meaningful drifts. Drift is introduced incrementally at multiple rates, where higher drift levels subsume lower ones.

Before drift, we compute ground-truth coverage statistics per environment to characterize task difficulty. The final dataset includes the initial resource state, persona query streams, relevance judgments, train--test splits, pre-computed embeddings, and drift event specifications. All stochastic components are seed-controlled to ensure reproducibility.

This construction enables systematic evaluation of routing policies under controlled resource fragmentation, partial observability, and increasing degrees of non-stationarity.

\paragraph{Tech Startup (cross-corpus) construction.} The Tech Startup benchmark is built outside the BEIR pipeline, by combining five distinct real corpora into a single five-resource environment template that mirrors a typical workplace stack. The five resources are sourced as follows: \emph{emails} from Enron \citep{Klimt2004enron}, \emph{chat channels} from the Alpaca instruction-tuning corpus \citep{taori2023alpaca} (response = document, instruction-derived rephrasing = query), \emph{issue tracker} from SWE-bench's GitHub issues \citep{jimenez2024swebench}, \emph{wiki pages} from HotpotQA Wikipedia paragraphs \citep{yang2018hotpotqa}, and \emph{code repository} from CodeSearchNet \citep{husain2019codesearchnet}. We assemble five environments using a connected-component partition over (document, query) pairs to ensure that no document or query appears in more than one environment.

\paragraph{Query rewriting and leakage control.} Each query is individually rewritten by an LLM rewriter under explicit constraints: (i) reference document \emph{content} rather than the originating application, (ii) avoid surface words that name resources (``email'', ``Slack'', ``issue'', ``ticket'', ``commit'', ``pull request'', ``wiki'', etc.), and (iii) read as natural questions a colleague would ask. After rewriting we apply a regex-based blocklist filter that rejects any query still containing resource-revealing tokens; matched queries are re-rewritten until the filter passes. We do not claim that residual style or length artifacts are eliminated, since the underlying corpora differ in discourse type by construction (cross-corpus heterogeneity is the property we want to measure); however, the same artifacts are equally available to every baseline, and the cross-corpus regime's natural difficulty is what makes the supervised classifier (Learn2Rank) competitive in Table~\ref{tab:exp1_main}. Every retained query passes a semantic-retrieval quality gate that requires its ground-truth document to appear in the top-$5$ of dense retrieval (all-MiniLM-L6-v2) over the environment's documents. The filter removes $4.0\%$ of candidate queries; the surviving $1{,}127$ queries form the released benchmark.

\section{Trajectory Generation and Citation Statistics}
\label{app:trajectories}

To isolate routing differences from stochastic exploration variance, we pre-compute primary-agent trajectories once per environment and reuse them across \method, the bandit baselines, and all ablations. This protocol fixes downstream interaction evidence so that performance differences reflect how each method interprets and updates from the same retrieval traces, rather than differences in agent exploration. Trajectories are generated with a production-grade tool-using agent (GPT-5-mini\footnote{\url{https://platform.openai.com/docs/models/gpt-5-mini}}) configured with the structured \texttt{search\_resources} and \texttt{synthesize\_answer} tools described in Section~\ref{sec:method_citations}, with a per-rollout step budget of $5$ and a context length of $16$K tokens.

\paragraph{Citation rate and accuracy.} For each trajectory we measure the citation rate $\Pr_t[\mathcal{C}_t \neq \emptyset]$ and the citation accuracy $\mathbb{E}_t[\,|\rho(\mathcal{C}_t) \cap \bar{\mathcal{R}}_t| / |\rho(\mathcal{C}_t)| \mid \mathcal{C}_t \neq \emptyset\,]$. On FEVER, citation rates are high but vary across environments (mean $85.7\%$, median $90.1\%$) while citation accuracy is stable (mean $71.3\%$, std.\ $4.9\%$). On NFCorpus, both rate (mean $74.3\%$) and accuracy (mean $73.1\%$) are stable. On Tech Startup, citation rate is consistently high (mean $\sim$$95\%$) and accuracy is high (mean $\sim$$73\%$) across the five resource types. Low citation rates yield fewer memory updates rather than noisier ones, supporting the observation that the citation gate trades coverage for precision, and motivating the noise-robustness analysis in Section~\ref{sec:results}.

\section{Training and Inference Hyperparameters}
\label{app:hyperparams}

\paragraph{Routing policy.}
\method is implemented as a LoRA-adapted routing policy on top of a Qwen3-8B base model and trained using Group Relative Policy Optimization (GRPO). LoRA adapters use rank $16$ and are optimized with a learning rate of $5\times10^{-6}$ and gradient accumulation over $4$ steps. Meta-training is performed for $25$ GRPO iterations; each iteration samples $8$ queries and generates $4$ rollouts per query (GRPO group size for advantage normalization). Queries whose group has zero-variance reward (e.g., all rollouts succeed or all fail) are filtered out by GRPO's advantage normalization and contribute no gradient signal.

\paragraph{Rollout and decoding settings.}
Rollouts are generated with a maximum of 5 interaction turns and a context length of up to 16K tokens. Decoding uses temperature $0.7$ and nucleus sampling with $p=0.95$.

\paragraph{Memory and retrieval.}
Unless otherwise stated, the routing agent maintains a per-resource memory budget of $M=100$ documents. At each routing decision, the agent retrieves the top $25$ memory entries via semantic search to constrain context length; this retrieval limit is independent of the per-resource memory budget. Memory updates are enabled incrementally at test time and are gated on agent-produced citations.

\paragraph{Embedding model.}
All embedding-based components across methods use the same frozen sentence encoder, sentence-transformers/all-MiniLM-L6-v2.

\paragraph{Bandit baseline.}
LinUCB-D is a discounted linear UCB bandit with one arm per resource. Each arm is represented by the centroid of the embeddings of up to $100$ documents sampled from the resource, and the context for query $q$ and resource $r$ is a fixed random projection (dimension $32$) of the elementwise product of their embeddings, concatenated with their cosine similarity and a bias term ($d{=}34$). It selects the top-$B$ resources by UCB score with exploration $\alpha{=}1.0$, ridge regularization $\lambda{=}1.0$ and discount $\gamma{=}0.999$. In the closed loop, after the agent acts on its recommendation, each selected resource receives reward $1$ if the agent cited a document from it and $0$ otherwise.

\section{Additional Ablations}
\label{app:additional_ablations}

This appendix collects further \method ablations that do not fit in the main paper.

\subsection{Closed-loop validation}
\label{app:closed_loop}

Table~\ref{tab:exp1_main} is measured closed-loop. Here we compare it with the offline protocol, which replays the same pre-computed primary-agent trajectories for every method so that routing differences are isolated from the agent's exploration variance. Offline, \method's memory is updated from the citations in those pre-computed trajectories; closed-loop, it is updated from the citations the agent emits after acting on \method's own recommendations. Table~\ref{tab:closed_loop} reports both.

\begin{table}[h]
    \centering
    \begin{threeparttable}
        \caption{Closed-loop vs.\ offline \method.}
        \label{tab:closed_loop}
        \small
        \begin{tabular}{lcccccc}
        \toprule
         & \multicolumn{3}{c}{Offline} & \multicolumn{3}{c}{Closed-loop} \\
         \cmidrule(lr){2-4} \cmidrule(lr){5-7}
        Dataset & $F1_{r}$ & $F1_{c}$ & Saved\,\% & $F1_{r}$ & $F1_{c}$ & Saved\,\% \\
        \midrule
        NFCorpus & $50.69$ & $20.05$ & $64.8$ & $47.67$ & $17.10$ & $61.2$ \\
        FEVER & $67.94$ & $59.70$ & $79.3$ & $64.72$ & $50.88$ & $73.7$ \\
        Tech Startup & $83.39$ & $77.66$ & $55.8$ & $86.60$ & $80.78$ & $60.4$ \\
        \bottomrule
        \end{tabular}
        \begin{tablenotes}
            \small
            \item \revred{$F1_{r}$: routing F1 at budget $B$ (Appendix~\ref{app:metrics}). $F1_{c}$: the agent's citation F1 against the relevance judgments. Saved\,\%: reduction in the agent's search calls relative to no routing.} Closed-loop values are \method's row of Table~\ref{tab:exp1_main}.
        \end{tablenotes}
    \end{threeparttable}
\end{table}

Closed-loop routing F1 is within $3.3$ points of offline on every dataset: $3.0$ and $3.2$ points lower on NFCorpus and FEVER, and $3.2$ points higher on Tech Startup. Citation F1 and Saved\,\% move in the same direction, with the largest change in FEVER citation F1 ($59.70$ to $50.88$). \method's advantage over the baselines in Table~\ref{tab:exp1_main} therefore does not depend on replaying pre-computed trajectories.

\paragraph{Scope and fairness.} All routers in Table~\ref{tab:exp1_main} run under the same closed-loop protocol: the same primary agent, query stream, routing budget and hard-routed replay. Static routers have no online state; LinUCB-D and \method are updated from the agent's citations as the stream is processed. We do not report latency.

\subsection{Robustness to citation noise}
\label{app:citation_noise}

A key claim of \method is that gating memory updates on agent citations rejects retrieval noise without requiring labels. We stress-test this gate on Tech Startup by perturbing the cited-document field of every trajectory before it reaches the router: \emph{drops} remove a fraction of citations entirely; \emph{flips} replace a fraction's resource provenance with a uniformly chosen incorrect resource. We additionally compare against an \emph{ungated} variant that updates memory from all retrieved documents per trajectory rather than only the cited ones. Each rate is run with a fixed seed. Table~\ref{tab:citation_noise} reports the outcome.

\begin{table}[h]
    \centering
    \begin{threeparttable}
        \caption{Citation-noise stress test on Tech Startup.}
        \label{tab:citation_noise}
        \small
        \begin{tabular}{lccccc}
            \toprule
            Variant & 0\% & 10\% & 25\% & 50\% & ungated \\
            \midrule
            \method (drop)  & 81.18 & 85.22 & 82.64 & 86.71 & --\\
            \method (flip)  & 81.98 & 64.48 & 46.84 & 36.48 & --\\
            no citation gate & --& --& --& --& 84.50 \\
            \bottomrule
        \end{tabular}
        \begin{tablenotes}
            \small
            \item $B{=}1$; values are F$_1$. \emph{drop} removes citations; \emph{flip} rewrites a citation's resource provenance to an incorrect resource; \emph{ungated} updates memory from all retrieved documents.
        \end{tablenotes}
    \end{threeparttable}
\end{table}

Drops barely move F$_1$ (range $81$--$87$): the gate already filters what dropping would discard, so the retained citations remain a high-precision signal. Flips produce a clean monotonic decline ($82 \!\to\! 64 \!\to\! 47 \!\to\! 36$), graceful but proportional degradation when the gate's accuracy itself is corrupted. The ungated baseline ($84.50$) sits slightly above gated drop $0\%$ ($81.18$): on Tech Startup, where the five resources come from semantically disjoint corpora, retrievals are already high-precision and the gate's filtering removes mildly useful corroborating evidence. The gate's value is therefore robustness: the contrast between a flat drop curve and a steep flip curve isolates the failure mode \method can recover from from the one it cannot, plus reduced storage from cited-only memory updates.

\subsection{Memory Design (mem0 comparison)}
\label{app:ablation_memory_design}

We compare \method's environment-structured memory against a generic semantic-memory baseline implemented with mem0\footnote{\url{https://mem0.ai/}}. The mem0 baseline stores past query--resource trajectories as text, embeds them with a sentence-transformer model, and retrieves semantically similar trajectories by vector similarity; candidate resources are then extracted from the retrieved trajectories. \method's memory instead represents resources as persistent entities and records interaction outcomes per resource, enabling reasoning about resource-specific reliability over time. Both variants share the same base model, resource representations, routing budget, and training procedure; only the memory structure differs.

\begin{table}[h]
    \centering
    \begin{threeparttable}
        \caption{Generic vs.\ environment-structured memory.}
        \label{tab:ablation_memory_design}
        \small
        \begin{tabular}{lcccccc}
         & \multicolumn{3}{c}{NFCorpus} & \multicolumn{3}{c}{FEVER} \\
         \toprule
        \textbf{Router} & $R_B \uparrow$ & $P_B \uparrow$ & $F1_B \uparrow$ & $R_B \uparrow$ & $P_B \uparrow$ & $F1_B \uparrow$\\
        \midrule
        w/ mem$0$ & 51.04 & 44.41 & 47.49 & 75.32 & 59.35 & 66.39 \\
        w/ env. mem. & \textbf{53.93} & \textbf{47.82} & \textbf{50.69} & \textbf{77.62} & \textbf{60.41} & \textbf{67.94} \\
        \bottomrule
        \end{tabular}
        \begin{tablenotes}
            \small
            \item Comparison between generic semantic memory (mem0) and \method's environment-structured memory. Results are averaged over $k$ folds of environment-level test splits.
        \end{tablenotes}
    \end{threeparttable}
\end{table}

Table~\ref{tab:ablation_memory_design} shows that environment-structured memory outperforms mem0 on both datasets. mem0's flat semantic store mixes exploratory tool calls, intermediate reasoning, and retrieved snippets with highly variable signal-to-noise, so credit assignment is diffuse. \method instead maintains resource-specific sketches updated only when the agent produces citations, yielding a higher-precision feedback channel and a stable per-resource representation of ``what this resource is useful for.''

\subsection{Memory Size Sensitivity}
\label{app:ablation_memory_size}

We study \method's sensitivity to per-resource memory capacity by varying the maximum number of retained documents from $5$ to $100$ while holding the routing policy, reward, and training procedure fixed.

\begin{figure}[h]
    \centering
    \includegraphics[width=0.85\linewidth]{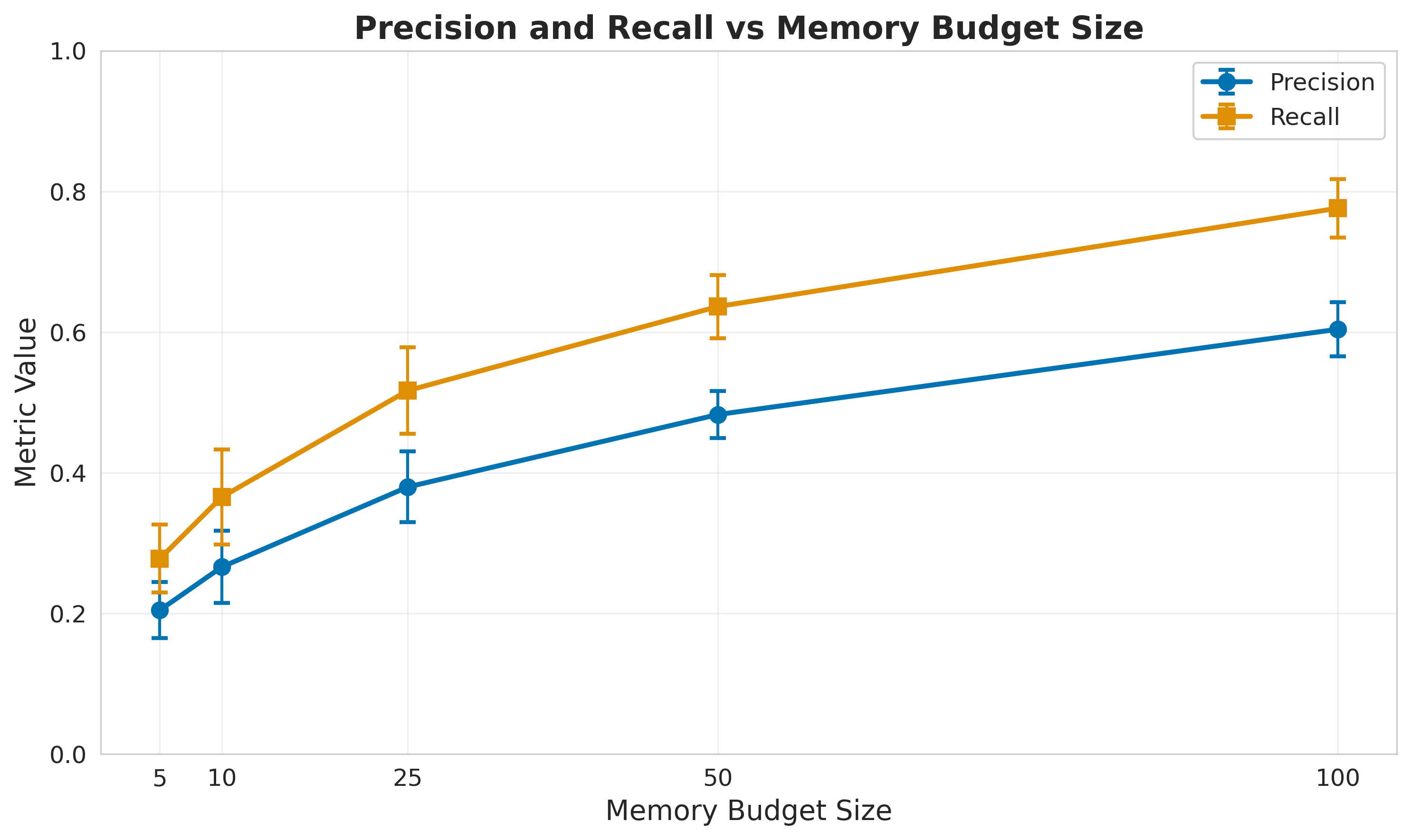}
    \caption{Precision and Recall at $B$ as a function of the per-resource memory budget. Gains diminish beyond moderate memory sizes, indicating that most useful routing signal is captured with limited historical context.}
    \label{fig:memory_size}
\end{figure}

Figure~\ref{fig:memory_size} shows that both precision and recall improve with memory capacity, but with diminishing returns beyond moderate sizes, supporting the practicality of \method under realistic memory budgets.

\section{Metric Definitions}
\label{app:metrics}

This appendix provides formal definitions of the evaluation metrics used in the experimental section. Following standard practice in federated search and collection selection
\citep{french2000metrics,shokouhi2011federated},
we assign each resource $r\in\mathcal{R}$ a nonnegative relevance merit for request $q_t$ based on document-level evidence:
\begin{equation}
m_t(r) \triangleq \sum_{d \in \mathrm{Top}L(r,q_t)} g(\mathrm{rel}(q_t,d)),
\end{equation}
where $\mathrm{Top}L(r,q_t)$ denotes the top-$L$ retrieved items from resource $r$,
$\mathrm{rel}(q_t,d)$ is a judged or verifier-derived relevance label, and $g(\cdot)$
is a fixed gain function (e.g., $g(\ell)=2^\ell-1$ for graded relevance).
The merit function is used only to rank relevant resources when constructing $\mathcal{R}^\star(q_t)$.

\paragraph{Recall at $B$ ($R_B$)}
Given a routing budget $B$, Recall at $B$ ($R_B$) measures the fraction of relevant resources recovered by the routed set:
\begin{equation}
R_B(q_t) \triangleq
\frac{|a_t \cap \mathcal{R}^\star(q_t)|}
{|\mathcal{R}^\star(q_t)|},
\end{equation}
where $a_t \subseteq \mathcal{R}$ denotes the set of resources selected for query $q_t$ (truncated to size $B$ when necessary), and $\mathcal{R}^\star(q_t)$ is the set of up to $B$ ground-truth relevant resources for $q_t$ with the highest merit. This metric evaluates how effectively the router identifies relevant resources under a fixed selection budget and is standard in federated search evaluation \citep{french2000metrics,shokouhi2011federated}.

\paragraph{Precision at $B$ ($P_B$)}
measures the fraction of selected resources that are relevant:
\begin{equation}
P_B(q_t) \triangleq
\frac{|a_t \cap \mathcal{R}^\star(q_t)|}{|a_t|},
\end{equation}
with $P_B(q_t)=0$ when $|a_t|=0$. This metric quantifies over-selection and complements recall-oriented measures by penalizing the inclusion of irrelevant resources.

\paragraph{$F1$ at $B$ ($F1_B$)}
summarizes the trade-off between selecting relevant resources and avoiding irrelevant ones, we report the harmonic mean of $P_B$ and $R_B$:
\begin{equation}
F1_B(q_t) \triangleq 
\frac{2 \cdot P_B(q_t) \cdot R_B(q_t)}
{P_B(q_t) + R_B(q_t)},
\end{equation}
with $F1_B(q_t)=0$ when both $P_B(q_t)$ and $R_B(q_t)$ are zero. $F1_B$ provides a single summary measure of routing quality that balances effectiveness and noise under a fixed resource-selection budget.

\end{document}